\documentclass[letterpaper, 10 pt, journal, twoside]{IEEEtran} 

\usepackage{graphics} 
\usepackage{graphicx}   
\usepackage{amsmath} 
\usepackage{amssymb}  
\usepackage{mathtools, cuted}
\usepackage{interval}
\usepackage{nicefrac}
\usepackage[noadjust]{cite}
\usepackage{breqn}
\usepackage{color, colortbl}
\usepackage{derivative}
\usepackage{siunitx}
\usepackage{multirow}

\renewcommand{\vec}[1]{\mathbf{\boldsymbol{#1}}}

\providecommand{\R}{\ensuremath \mathbb{R}}
\providecommand{\N}{\ensuremath \mathbb{N}}

\newcommand{\ie}{\textit{i}.\textit{e}., }

\usepackage[dvipsnames]{xcolor}

\title{Practice Makes Perfect: an iterative approach to achieve precise tracking for legged robots}

\author{Jing Cheng, Yasser G. Alqaham, Amit K. Sanyal, and Zhenyu Gan
\thanks{All authors are with the Department of Mechanical and Aerospace Engineering, Syracuse University, Syracuse, NY 13244 \texttt{\{jcheng13, ygalqaha, aksanyal, zgan02\}@syr.edu}.}
\thanks{This work was supported by a startup fund from the Syracuse University.}
}

\begin{document}

\maketitle

\begin{abstract}
Precise trajectory tracking for legged robots can be challenging due to their high degrees of freedom, unmodeled nonlinear dynamics, or random disturbances from the environment.
A commonly adopted solution to overcome these challenges is to use optimization-based algorithms and approximate the system with a simplified, reduced-order model.
Additionally, deep neural networks are becoming a more promising option for achieving agile and robust legged locomotion.
These approaches, however, either require large amounts of onboard calculations or the collection of millions of data points from a single robot.
To address these problems and improve tracking performance, this paper proposes a method based on iterative learning control. This method lets a robot learn from its own mistakes by exploiting the repetitive nature of legged locomotion within only a few trials.
Then, a torque library is created as a lookup table so that the robot does not need to repeat calculations or learn the same skill over and over again.
This process resembles how animals learn their muscle memories in nature.
The proposed method is tested on the A1 robot in a simulated environment, and it allows the robot to pronk at different speeds while precisely following the reference trajectories without heavy calculations.  
\end{abstract}


\section{Introduction}

\IEEEPARstart{L}{egged} robots such as the bipedal humanoid robot \emph{Atlas} and the quadrupedal robot \emph{Spot} from Boston Dynamics \cite{guizzo2019leaps} are drawing the attention of the media worldwide due to their amazing agility and mobility in complex environments.
These robots are ideal platforms to replace human workers from labor-intensive and repetitive tasks in manufacturing lines or on construction sites.
They are also suitable alternatives for jobs that may present a risk of injury, such as search and rescue missions, firefighting, and inspecting nuclear plants and sites where toxic waste may be present.
However, their highly dynamic motions come with costs associated with mechanical design, modeling, and software control. 
In order to enable animal-like reflexes, these robots usually have a high degree of freedom (DOF) in their auxiliary bodies, such as legs and arms.
These added bodies not only increase the difficulties in the joint mechanism design but also introduce new uncertainties in measuring associated parameters such as friction coefficients across joints or reflected inertia proprieties of the distal parts of the limbs.
Therefore modeling the system in every detail is challenging and the resulting equations of motion (EOM) are usually complex.
Additionally, to perform skills and maneuvers, a control algorithm has to be implemented to constantly generate reference trajectories for the joints and to stabilize the motion of the whole body.
Through this process, the EOM needs to be used repeatedly to ensure the generated control policy is realistic and realizable in a physical environment.
However, evaluating these equations in real-time requires fast symbolic and numerical calculations onboard and consumes a large amount of computational resources.
%
\begin{figure}[tbp]
\centering
\includegraphics[width=1\columnwidth]{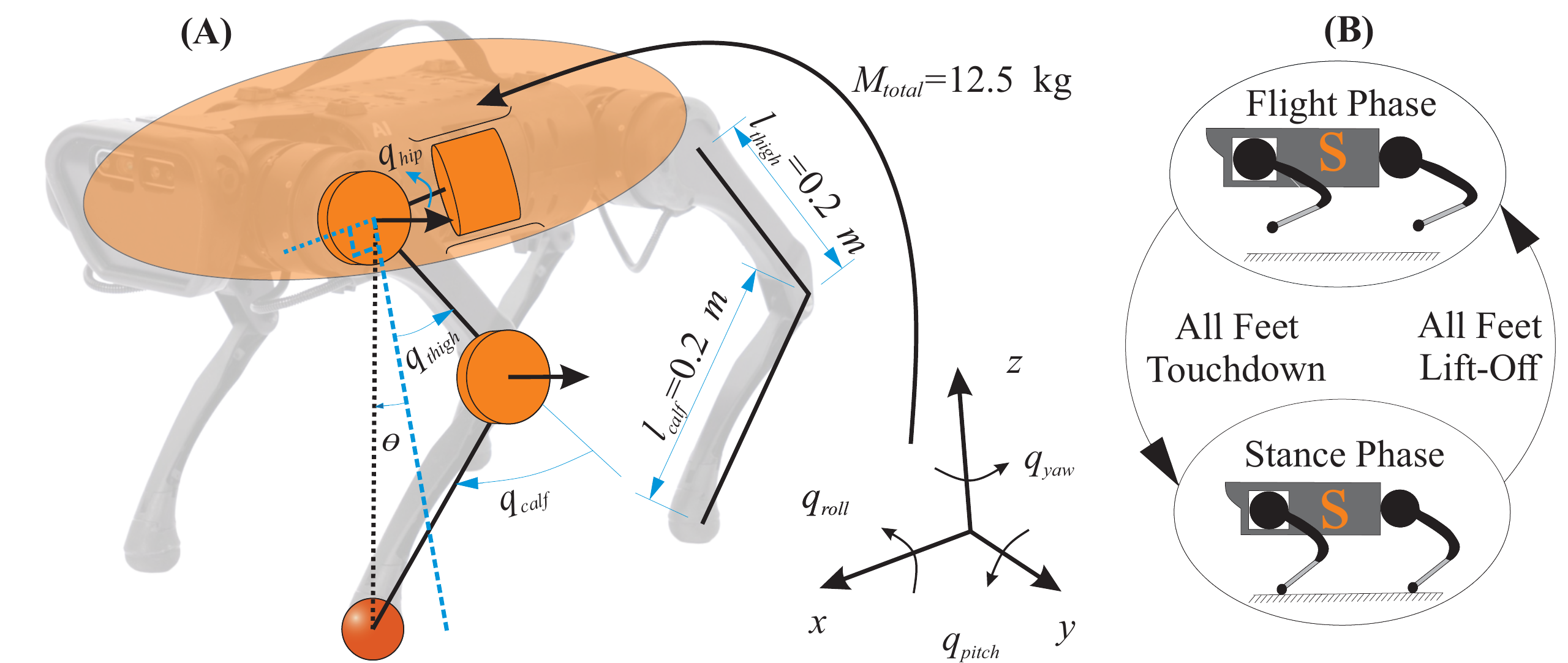}
\caption[Model used in this study]{The configuration of quadrupedal robot A1 from Unitree Robotics and the state machine of the pronking gait.}
\label{fig:statemachine}
\vspace{-2mm}
\end{figure}

Two primary approaches have been adopted by roboticists in the past years to design locomotion controllers and resolve the aforementioned problems.
The first approach is to ignore the detailed motions of segmented bodies and to focus on the centroidal dynamics of the torso by using simplified models such as the linear pendulum model (LPM) \cite{kajita20013d} or the single rigid body model (SRB).
For example, with the reduced order SRB model, the MIT mini-cheetah robot can plan the desired trajectories and ground reaction forces (GRF) in real-time using model predictive control (MPC) and solving a convex quadratic programming (QP) problem \cite{mitcheetah, Katz_2019}.
Thereafter, a whole body model with physical constraints is used to solve the inverse dynamics problem to ensure the desired motion is achievable on hardware.
With simplified models, the algorithms can be implemented in the onboard computers.
In general, they work well in controlling the robot's motion with different footfall patterns.
However, this approach has no guaranteed stability, and the performance of the MPC methods depends on the convergence of the QP problem.
In addition, it is costly to run such algorithms since the simplified models are not accurate representations of real robots.
As a result, the time horizon of the MPC is usually short and the algorithm needs to be reevaluated at a rate of more than \SI{200}{\hertz} \cite{mitcheetah}.
It is also important to note that the joint tracking performance depends on the accuracy of the whole body model, which often fails to capture the full dynamics of the robot.

On the other hand, \emph{machine learning} (ML) approach has been widely embraced by the robotics community in the past decade, it serves as a versatile and lightweight alternative for the problem of locomotion.
For example, the bipedal Cassie robot at Oregon State University can climb up/down stairs without the external help of a vision system \cite{siekmann2021blind}.
Additionally, ANYmal from ETH demonstrated a highly robust locomotion controller trained using reinforcement learning in simulation over challenging terrain \cite{lee2020learning}.
Nevertheless, this approach is known for its data-hungry nature and a deep neural network (DNN) needs to be trained on high-performance clusters for days or even months.
Furthermore, it suffers from a simulation-to-reality transition gap since most of the data used for DNN training is gathered from simulations, while only a small portion of data can be collected from hardware \cite{9308468}.

Legged animals in nature, however, are not solving complicated math problems to decide how to coordinate the motions of their bodies and limbs when moving around.
Neither do they perform the same task millions of times to gather ``big data" to train their brains and muscles when they learn a new skill.
In fact, many mammals, including deer fawns, can walk after only a few trials within hours of their birth \cite{jackson1972activity}. 
In this paper, we propose a bioinspired controller design framework that does not rely on heavy calculations of the model dynamics when the controller is running, nor does it need a large amount of data and computational resources power to achieve accurate and stable motions.
By leveraging the idea of \emph{iterative learning control} (ILC) \cite{kasac2008passive, ahn2007iterative} and exploiting the repetitive nature of the locomotion tasks, our proposed controller can learn from its own mistakes regardless of the unmodeled dynamics within only a few seconds.
The idea of ILC was first proposed in the 1980s, and it was widely adopted in the development of industrial manipulators in manufacturing processes, such as robot arms in car-body paint shops \cite{bristow2006survey}.
This work rejuvenates the idea by incorporating it into the legged locomotion controller to learn unmodeled robot dynamics rapidly.
The proposed controller is implemented on a quadrupedal robot called A1 (Fig. 1A) from Unitree Robotics (as introduced in Section \ref{sec:ModelDescription}).
The remainder of Section \ref{sec:methods} demonstrates the detailed structure of the ILC.
The performance of the proposed approach for pronking gait is shown in Section \ref{sec:results}, and Section \ref{sec:conclusion} concludes the paper.

\section{Methods}
\label{sec:methods}
This section introduces the notation, the dynamic model of the A1 robot, and the pronking gait library generated through trajectory optimizations. 
The formulation and implementation of the ILC policy are also explained in detail. 

\subsection{Dynamic Model}
\label{sec:ModelDescription}
In this paper, we focus on quadrupedal \emph{pronking} gait only (Fig. 1B), in which all four legs move in a fully synchronized fashion with two distinct phases: the \emph{flight phase} and the \emph{stance phase}, assuming that all four feet strike and lift off the ground simultaneously.
In order to obtain a consistent set of equations of motion for both phases, we adopt a floating-base model using the generalized coordinates as follows:
%
\begin{equation}
\label{eq:configuration variables}
   \vec{q} \mathrel{\mathop:}= \left[\vec{q}_{B}^T, \vec{q}_{L}^T \right]^T, 
\end{equation}
%
where $\vec{q}_{B}$ represents the 3-dimensional translational and rotational motions of the rigid torso in the inertial frame, and $\vec{q}_{L}$ refers to the rotational motions of auxiliary bodies used for legged locomotion measured in the body coordinate frames.
As shown in Fig. 1A, on the A1 robot, each limb has three revolute joints measured by relative angles, namely, hip joint ($q_{\text{hip}}$), thigh joint ($q_{\text{thigh}}$), and calf joint ($q_{\text{calf}}$).
The first two joints are driven directly by brushless DC motors with a transmission ratio of 9.
The calf joints are teleoperated by another DC motor mounted on the thigh through four-bar linkages to further reduce the moments of inertia of the limbs.
There are in total 12 revolute joints which are listed in the order of front right (FR) leg, front left (FL) leg, rear right (RR) leg, and rear left (RL) leg in $\vec{q}_{L}$ respectively.
The equations of motion of the A1 robot can be derived from the Euler-Lagrange equation:
\begin{equation}
\label{eq:EOM}
   \operatorname{\vec{M}}(\vec{q})\ddot{\vec{q}} + \operatorname{\vec{H}}(\vec{q},\dot{\vec{q}})\dot{\vec{q}} + \operatorname{\vec{G}}(\vec{q}) = \vec{S}\vec{\tau}  + \vec{J}^T\vec{\lambda} ,
\end{equation}
where $\operatorname{\vec{M}}(\vec{q})\in\R^{n_{R} \times n_{R}}$ is the mass matrix; $\operatorname{\vec{H}}(\vec{q},\dot{\vec{q}})\in\R^{n_{R}}$ is the vector consisting of Coriolis forces and centrifugal forces;
and $\operatorname{\vec{G}}(\vec{q})$ is the vector of gravitational forces. 
$\vec{\tau} \in\R^{n_{L}}$ denotes the vector of motor torques acting on the revolute joints and $\vec{S} = \left[\vec{0}_{n_B \times n_L}; \vec{I}_{n_L \times n_L}\right]$ is the selection matrix which contains a zero submatrix and an identity submatrix to distribute the motor torques to each individual joint. $n_{R}$, $n_{L}$, and $n_{B}$ are 18, 12, and 6, respectively.  

In the flight phase, the system has the structure of a kinematic tree and the GRF vector $\vec{\lambda}$ is zero. In the stance phase, however, the robot's feet are in contact with the ground, and the GRFs are denoted as $\vec{\lambda}_l \in \R^{3}$ ($l \in \{\text{FR, FL, RR, RL\}}$).
For simplicity, we assume the feet in stance phase never slip on the ground and the following holonomic constraints are always satisfied:
\begin{align}
       \vec{c}_l = \vec{g}_l\left(\vec{q}\right), \quad
       \dot{\vec{c}}_l  = \vec{J}^T_l\dot{\vec{q}} = \vec{0}, \quad
       \ddot{\vec{c}}_l &= \vec{J}^T_l\ddot{\vec{q}} + \vec{\sigma}_l = \vec{0}, \label{eq:cv}
\end{align}
where $\vec{g}_l$ denotes the set of forward kinematic equations; $\vec{c}_l \in\R^{3}$ is the foot location in the inertial frame; $\vec{J}_l \mathrel{\mathop:}=  \pdv{\vec{g}_l}{\vec{q}} \in\R^{3\times{n}_{R}}$ is the corresponding Jacobian matrix of the foot contact and $\vec{\sigma}_l \mathrel{\mathop:}=  \vec{\dot{J}}_l^T\dot{\vec{q}}$ is the bias acceleration term \cite[Chapter~2]{Rigidbody_ROY}.
We can further stack the last equation in (\ref{eq:cv}) for all four legs to close the loop of the kinematic tree by enforcing:
\begin{equation}
\label{eq:cvall}
   \ddot{\vec{c}} = \vec{J}^T\ddot{\vec{q}} + \vec{\sigma} = \vec{0} \in\R^{12},
\end{equation}
By combining (\ref{eq:EOM}) and (\ref{eq:cvall}), we obtain the following differential-algebraic equations for the stance phase dynamics:
%
\begin{equation}
\label{eq:DAE}
    \begin{bmatrix}
           \operatorname{\vec{M}}(\vec{q}) & -\vec{J}^T \\
           \vec{J}^T &  \vec{0}
    \end{bmatrix}
     \begin{bmatrix}
           \ddot{\vec{q}} \\
           \vec{\lambda}
    \end{bmatrix}
    = \begin{bmatrix}
           \vec{S}\vec{\tau} - \operatorname{\vec{H}}(\vec{q},\dot{\vec{q}})\dot{\vec{q}} - \operatorname{\vec{G}}(\vec{q})  \\
           -\vec{\sigma}
    \end{bmatrix}.
\end{equation}
%
\subsection{Trajectory Generation}
\label{sec:TrajGeneration}
In the proposed controller, a \emph{gait library} (GL) of optimal solutions is generated offline and used to provide reference trajectories for the controller design of the A1 robot. The gait generation task is formulated as a hybrid trajectory optimization problem, and it is transcribed to a nonlinear program by the open-source FROST framework \cite{hereid2017frost} using the direct collocation method with Hermite-Simpson integration scheme.

The solutions are optimized in FROST to attain the pronking gait with a fixed apex height of $0.34$ \si{\meter} and varying average speeds while minimizing the following cost function:
\begin{equation}
\label{eq:cost function}
   \operatorname J=\int_{t_0}^{t_f}\vec{\tau}^T\vec{W}_{\tau}\vec{\tau}  +  \dot{\vec{q}}_L^T\vec{W}_{q}\dot{\vec{q}}_L\:d\xi 
\end{equation}
where $t_0$ and $t_f$ are the start and end times of the stride, and $\vec{W}_{\tau}$ and $\vec{W}_{q}$ are constant diagonal weight matrices that are used to reduce energy consumption and minimize leg passive swinging motions throughout the stride.
The following list of constraints is imposed:

\begin{itemize}
\item Average speed: $\bar{v} = \frac{q_{x}(t_f) - q_{x}(t_0)}{t_f - t_0}$;
\item Apex height: $ z_{max} = {\scriptstyle  0.34 \: m }$;
\item Dynamic constraints (\ref{eq:EOM}) ;
\item Holonomic constraints in stance (3) ;
\item Unilateral constraints in flight: $\vec{g}_l(\vec{q})> 0$ ;
\item Periodicity constraints: $\vec{q}(t_0) = \vec{q}(t_f)$, $q_{x}(t_0) \ne q_{x}(t_f)$;
\item Configuration limits: $\vec{q}_{min}\leq\vec{q}\leq\vec{q}_{max}$;
\item Velocity limits: $|\dot{\vec{q}}|\:\leq\dot{\vec{q}}_{max}$;
\item Torque limits: $|\vec{\tau}|\:\leq\vec{\tau}_{max}$;
\item Friction cone limits: $||\vec{\lambda}_t||_2 \:\leq \mu \: ||\vec{\lambda}_n||_2$;
\item Bilateral symmetry constraint.
\end{itemize}
In the above constraints, $q_{x}$ is the torso's $x$ position; $z_{max}$ is torso's apex height; $\mu$ is the friction coefficient of the ground contact; ($\vec{\lambda}_t$ and $\vec{\lambda}_n$ are tangent and normal components of the GRFs with respect to the contact surface).
The trajectory of each joint ($j \in \N_+, 1 \leq j \leq n_{L}$) from the optimization scheme is approximated by a B\'ezier polynomial with an order of $n_M$:
%
\begin{equation}
\label{eq:bezier ploynomials}
   h_{j}(s) = \sum_{i=0}^{n_M} \alpha_{j,i} \frac{n_M !}{i !(n_M-i) !} s^{i}(1-s)^{n_M-i}
\end{equation}
%
where $s \in [0,1]$ is the phase variable and $\alpha_{j,i}$ are constant B\'ezier coefficients.
$s = 0$ indicates the start of the phase and $1$ means the end of current phase.
As a result, each phase of an optimal solution can be fully represented by a numerical matrix $\vec{B}(\bar{v})  \in \R^{n_{L} \times \left(n_{M}+1\right)}$ parameterized by the average velocity $\bar{v}$ as a collection of coefficients $\alpha_{j,i}$ of the B\'ezier polynomials for all joints.

\subsection{Trajectory Tracking Controller Design}
\label{sec:control}
%
\begin{figure}[tbp]
\centering
\includegraphics[width=1\columnwidth]{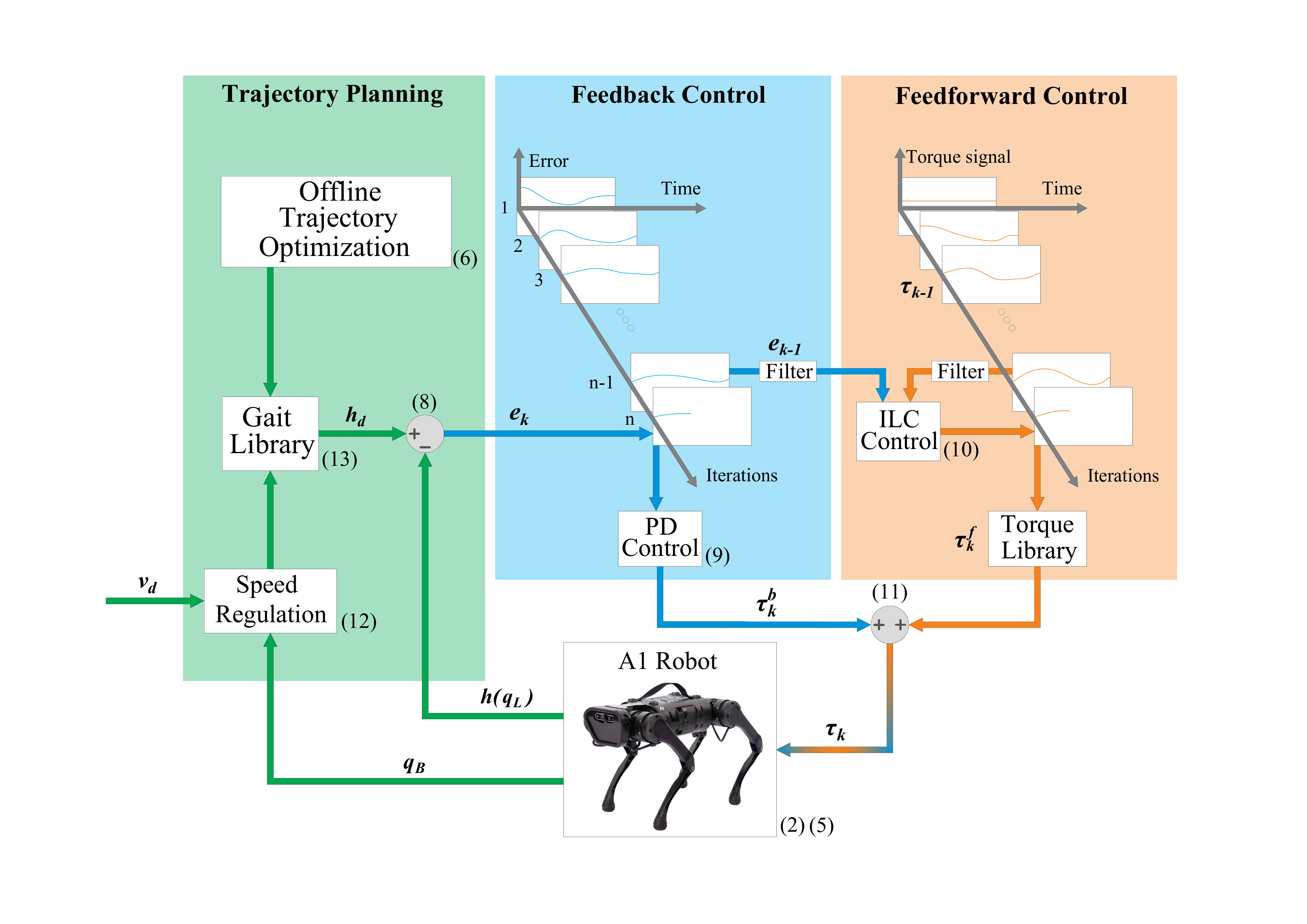}
\caption[Iterative learning control algorithm]{This figure illustrates the proposed control hierarchy. 
Trajectory planning, feedback control, and feedforward control are highlighted in green, blue, and orange regions respectfully.
The detailed calculations of all modules are listed with their corresponding equation numbers in the paper.}
\label{fig:controller structure}
\vspace{-2mm}
\end{figure}
%
The proposed trajectory tracking controller has two components.
The first component is a feedback term realized by a simple proportional-derivative (PD) controller solely based on the tracking error during the current iteration.
Additionally, the ILC acts as a feedforward term and continuously learns from the data obtained from the previous iterations to compensate for the system dynamics and external disturbance in the incoming tasks.
These two components are highlighted in blue and orange regions in Fig.~\ref{fig:controller structure} respectively.

\subsubsection{Feedback control}
suppose the desired trajectories at a given average speed $\bar{v}$ is denoted as $\vec{h}_{\text{d}}(\vec{B}(\bar{v}), s) \in\R^{n_{L}}$.
The trajectory tracking problem at $k$th iteration can be defined through $n_L$ virtual constraints to zero out the error functions $\vec{e}_k$ defined as follows:
%
\begin{equation}
\label{eq:virtualconstraint}
   \vec{e}_k(s) =\vec{h}_{\text{d}}(\vec{B}(v), s) - \vec{h}(\vec{q}_{L}, \frac{t_k}{T}),
\end{equation}
%
where $t_k$ is the time measured from the beginning of each phase during $k$th iteration and $T$ is the stride time. The feedback torque $\vec{\tau}^{b}_{k}(s)$ only depends on the tracking error and its derivative during the current iteration $k$ as calculated below:
%
\begin{equation}
\label{eq:feedback}
   \vec{\tau}^{b}_{k}(s) =K_P^{b} \, \vec{e}_k(s) + K_D^{b} \, \dot{\vec{e}}_k(s),
\end{equation}
%
in which $K_P^{b}$ and $K_D^{b}$ are the feedback proportional gain and derivative gain matrices respectively.

\subsubsection{Feedforward control}
the feedforward control usually can be implemented using inverse dynamics \cite[Chapter~9]{corke2011robotics}, however, the exact dynamic model of the robot is difficult to obtain, \ie the inverse dynamics method requires exact knowledge of the terms $\operatorname{\vec{M}}(\vec{q})$, $\operatorname{\vec{H}}(\vec{q},\dot{\vec{q}})$, $\operatorname{\vec{G}}(\vec{q})$, and $\vec{\lambda}$ in (\ref{eq:EOM}).
In order to learn the unmodeled dynamics and to improve the trajectory tracking performance for all joints in the robot, the ILC is implemented as a model-free approach by directly utilizing the recorded torques generated in the previous run $\vec{\tau}_{k-1}(s)$.
This is because for a repetitive task such as pronking motion, same reference trajectories are followed and similar tracking errors are observed during each stride.
To speed up the learning process, we look ahead in phase by a small amount $\delta s$ and precompensate for the incoming tracking errors using the errors generated from the previous iteration of $k-1$:
%
\begin{align}
\label{eq:feedforward}
   \vec{\tau}^{f}_{k}(s) = \vec{\tau}_{k-1}(s)  &+ K_P^{f} \, \vec{e}_{k-1}(s+\delta s) \\ \notag 
                            &+ K_D^{f} \, \dot{\vec{e}}_{k-1}(s+\delta s),
\end{align}
%
in the above equation, $K_P^{f}$ and $K_D^{f}$ are the feedforward proportional gain and derivative gain matrices.
The final desired torques sent to the robot are the combination of the feedback term (\ref{eq:feedback}) and feedforward term (\ref{eq:feedforward}) as illustrated in Fig.~\ref{fig:controller structure}.
%
\begin{equation}
\label{eq:desiredtorque}
   \vec{\tau}_{k}(s) = \vec{\tau}^{b}_{k}(s) + \vec{\tau}^{f}_{k}(s).
\end{equation}
%
$\vec{\tau}_{k}(s)$ is the vector of current torque inputs for all joints.

Similar control schemes have been successfully applied on many nonlinear systems in previous work.
The time and frequency domain convergence properties of iterative learning control are demonstrated in \cite{Norrl_f_2002}.
The proof of stability and convergence analysis for the ILC has been carried out for nonlinear systems which is similar to \cite{Horowitz_1993, Kasac_2008, Yin_2009}.
Due to the limitations of space, it is omitted here. 
%
\begin{figure*}[t!]
\centering
\includegraphics[width=2\columnwidth]{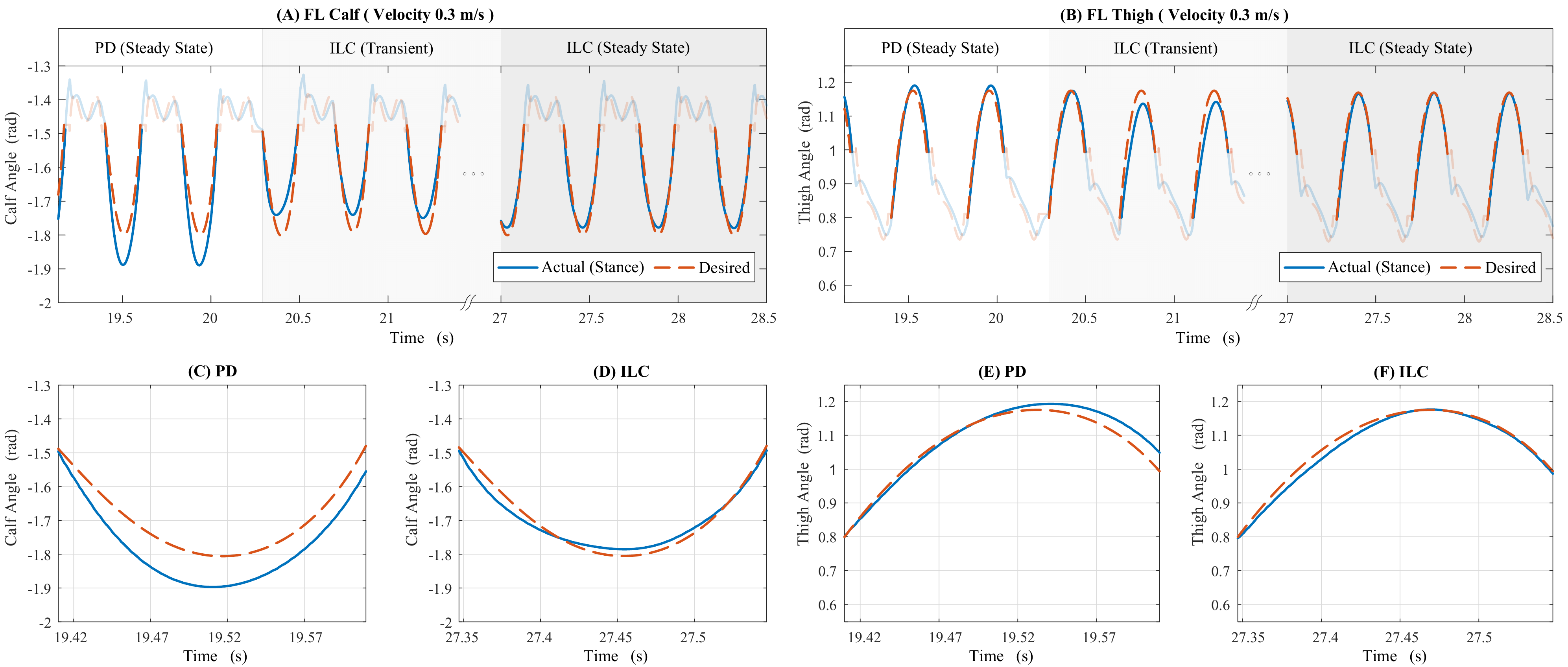}
\caption{The trajectory tracking performances of A1 robot are compared between the feedback control alone and the proposed control scheme with ILC in this figure. As shown in (A) and (B), the ILC is enabled at $t =20.3$ \si{\second} in both of the calf and thigh joints and the tracking errors in stance phases are greatly reduced in the steady state. In all figures the reference trajectories are shown in dash lines and actually trajectories are in solid lines.}
\label{fig:ILC_calf_thigh}
\vspace{-2mm}
\end{figure*}
%
\subsection{Longitudinal Speed Regulation}
When the A1 robot is pronking, the average speed of its torso in the longitudinal direction is regulated by an empirical foot-placement controller inspired by \cite[Chapter~3]{raibert1986legged}.
In this controller, the leg angle is modified using a feedback controller to reject the disturbance of the torso's velocity.
As for the A1 robot, the leg angle is defined as $\theta= q_{\text{thigh}} +\frac{1}{2} q_{\text{calf}}$.
This controller alters the leg angles $\theta$ in the sagittal plane prior to the ground contact based on the current torso's longitudinal velocity $v$:
%
\begin{equation}
    \label{eq:FPC}
    \theta_{d} =\theta + K_{\theta}\left(v-\bar{v}_{d}\right)
\end{equation}
%
where $K_{\theta}$ is a constant proportional gain. 
$\bar{v}_{d}$ is the desired velocity of the torso's centroid.
The corresponding joint angles are calculated through the inverse kinematics algorithm to track $\theta_{d}$.
Through the speed regulation process, the reference trajectories are continuously selected from the GL by interpolating the B\'ezier coefficients stored in $\vec{B}(\bar{v})$ using the current speed of the torso:
%
\begin{equation}
\label{eq:interpolation}
   \vec{B}(v) =\frac{\bar{v}_b - v}{\bar{v}_b- \bar{v}_a}\vec{B}(\bar{v}_a) + \frac{v - \bar{v}_a}{\bar{v}_b- \bar{v}_a}\vec{B}(\bar{v}_b)
\end{equation}
%
in which $\bar{v}_a$ and $\bar{v}_b$ are the two closest speeds selected from the GL and $\bar{v}_a \leq v < \bar{v}_b$.
A more detailed explanation of this choice can be found in \cite{da20162d, gong2019feedback}.

\subsection{Filter design}
\label{sec:filter}

The pronking gaits from the trajectory optimization are planned for 2 to 4 strides per second. A zero-phase filter is applied to the recorded data after each stride is finished. The purpose of that is to reduce the high-frequency noises in the error $\vec{e}_k(s)$ and torque $\vec{\tau}_k(s)$ signals and to speed up the learning process.
The filter is an infinite impulse response (IIR) digital filter, which is applied on the data through a forward pass followed by a backward pass, resulting in a signal with a zero phase shift \cite{kormylo1974two}.
This IIR filter with an order $j$ can be written as the following discrete transfer function:
%
\begin{equation}
\label{eq:IIR}
Y(z) = \frac{b_0+b_1 z^{-1} \cdots+b_{j} z^{-j}}{1+a_1 z^{-1} \cdots+a_{j} z^{-j}} X(z)
\end{equation}
%
where $a_{j}$ and $b_{j}$ are the feedback and feedforward filter coefficients; ${X(z)}$ and ${Y(z)}$ are the input and output signals.

Additionally, we use a first-order low-pass filter to smooth out the torso's velocity from the inertial measurement unit (IMU) sensor.
In order to overcome the drifts in the speed estimation, we use the leg odometry \cite{6094484} and replace the value of torso's velocity by assuming no foot slip and recalculating it using the inverse kinematics algorithm during stance phase.
\section{Results}
\label{sec:results}
To demonstrate the proposed control scheme is able to improve the trajectory tracking performance online, we implement the ILC in Robot Operating System (ROS) \cite{ROS} and simulate the A1 robot in a virtual environment in Gazebo \cite{koenig2004design}.
In general, the 3D motion of the robot can be achieved by including an additional speed regulation controller in the transverse direction \cite{gong2019feedback}.
However, the goal of this work is to illustrate the improvement of joint trajectory tracking performance.
For simplicity, the simulated robot is constrained to move only in the sagittal plane without the torso rotation.
The control policy is updated at a frequency of \SI{1}{\kilo\hertz}, and all calculations can be done in time on a desktop computer with Intel\textregistered \,  Core\textsuperscript{TM} i7 processor.
In this section, we demonstrate the comparisons of the tracking performance for two control schemes: feedback control alone, and the combined control strategy of feedback and ILC in (\ref{eq:desiredtorque}).
In total, $14$ optimal solutions are obtained from the trajectory optimization using the FROST framework.
The average speeds of these solutions are varied from \SI[per-mode=symbol]{-0.5}{\meter\per\second} to \SI[per-mode=symbol]{0.8}{\meter\per\second} incremented by \SI[per-mode=symbol]{0.1}{\meter\per\second}.
The coefficients of zero-phase filter are picked as ${\left\{a_0, a_1, a_2, a_3\right\}} = {\left\{1.00, -2.37, 1.93, -0.53\right\}}$ and ${\left\{b_0, b_1, b_2, b_3\right\}} = {\left\{0.003, 0.009, 0.009, 0.003\right\}}$.
In all simulations, we use the same the proportional gain matrices for $K_P^{b}$ and $K_P^{f}$ and the same derivative gain matrices for $K_D^{b}$ and $K_D^{f}$ for each leg. 
They are chosen to be $\big(\begin{smallmatrix} 35 & 0 & 0\\ 0 & 60 & 0\\ 0 & 0 & 90 \end{smallmatrix}\big)$ and $\big(\begin{smallmatrix} 1 & 0 & 0\\ 0 & 6 & 0\\0 & 0 & 2 \end{smallmatrix}\big)$.

\subsection{Improved Tracking Performance of ILC}
In the simulation, the robot first performs a startup procedure to go to a predefined posture and prepares itself for the first step of the pronking gait.
When the desired pronking speed command is received, the robot starts pronking and only the PD controller kicks in to track the reference trajectories from the GL.
As shown in Fig.~\ref{fig:ILC_calf_thigh}, the solid lines are the actual front left leg trajectories and the dash lines are the front left leg desired trajectories from the GL with an average speed of \SI[per-mode=symbol]{0.3}{\meter\per\second}.
When only the PD controller is engaged ($t<$ \SI{20.3}{\second}), the robot can in general track the reference trajectories.
However, in the steady state, large tracking errors are observed in both of the calf and thigh joints.
It can be seen that the largest tracking errors occur with the values of \SI{0.09}{\radian} roughly in the middle of stance phase for the calf joint (Fig.~\ref{fig:ILC_calf_thigh}C) and \SI{0.05}{\radian} when it is approaching the end of stance phase (Fig.~\ref{fig:ILC_calf_thigh}E) in thigh joint.
This is not surprising because the actuators, especially the ones that control calf joints, have to constantly compensate for the gravitational forces of the whole body during stance phase and all legs reach their shortest lengths with the smallest lever-arms in the mid-stance phase.

After that, we send out the command to start the ILC at \SI{20}{\second} and the ILC is not active until the new stance phase begins at \SI{20.3}{\second}.
The flight phase control remains the same, but the torques recorded from the previous stance phase are directly added in the stance phase control as a feedforward term according to (\ref{eq:feedforward}).  
As shown in Fig.~\ref{fig:ILC_calf_thigh}A and B, the directions of tracking errors in both joints are reversed, and it takes roughly 17 strides (\SI{7}{\second}) for ILC to reach to its steady state.
As we can see from Fig.~\ref{fig:ILC_calf_thigh}D, with the help of ILC, the tracking error in the calf joint is greatly reduced to \SI{0.02}{\radian}. 
As for the thigh joint, nearly perfect tracking is achieved after the mid-stance phase, and the largest tracking error is observed in the first half of the stance phase with a magnitude of only \SI{0.025}{\radian} (see Fig.~\ref{fig:ILC_calf_thigh}F).
Exactly the same simulation procedure is conducted for all reference trajectories stored in the GL.
As tabulated in Table~\ref{table:1}, we observe a similar trend in improvements in tracking behaviors for solutions with different speeds after the proposed ILC scheme is engaged.
On average, the tracking errors are reduced by 75.8\% in calf joints and 30.8\% in thigh joints after using the ILC.

\subsection{Learned Torque Library as the ``muscle memories"}
From the simulation results, with a given reference trajectory, the ILC demonstrates its effectiveness almost immediately by utilizing the data gathered from a few strides, and the whole learning process takes roughly less than \SI{10}{\second} to converge to the steady state with significantly reduced tracking errors.
As soon as the robot reaches its steady state, the feedforward joint torques $\vec{\tau}^{f}_{k}(s)$ can precisely compensate for the unmodeled dynamics and gravity, and they are no longer varying from one iteration to another, \ie $ \vec{\tau}^{f}_{k}(s) \approxeq \vec{\tau}_{k-1}(s)$.
Since the torque profiles for a given motor task are fixed, there is no need to learn the same skill every time when the skill is performed.
We can save the torque profiles of all motors and link them with each task as a lookup table.

\begin{figure}[tb]
\centering
\vspace{-10mm}
\includegraphics[width=1\columnwidth]{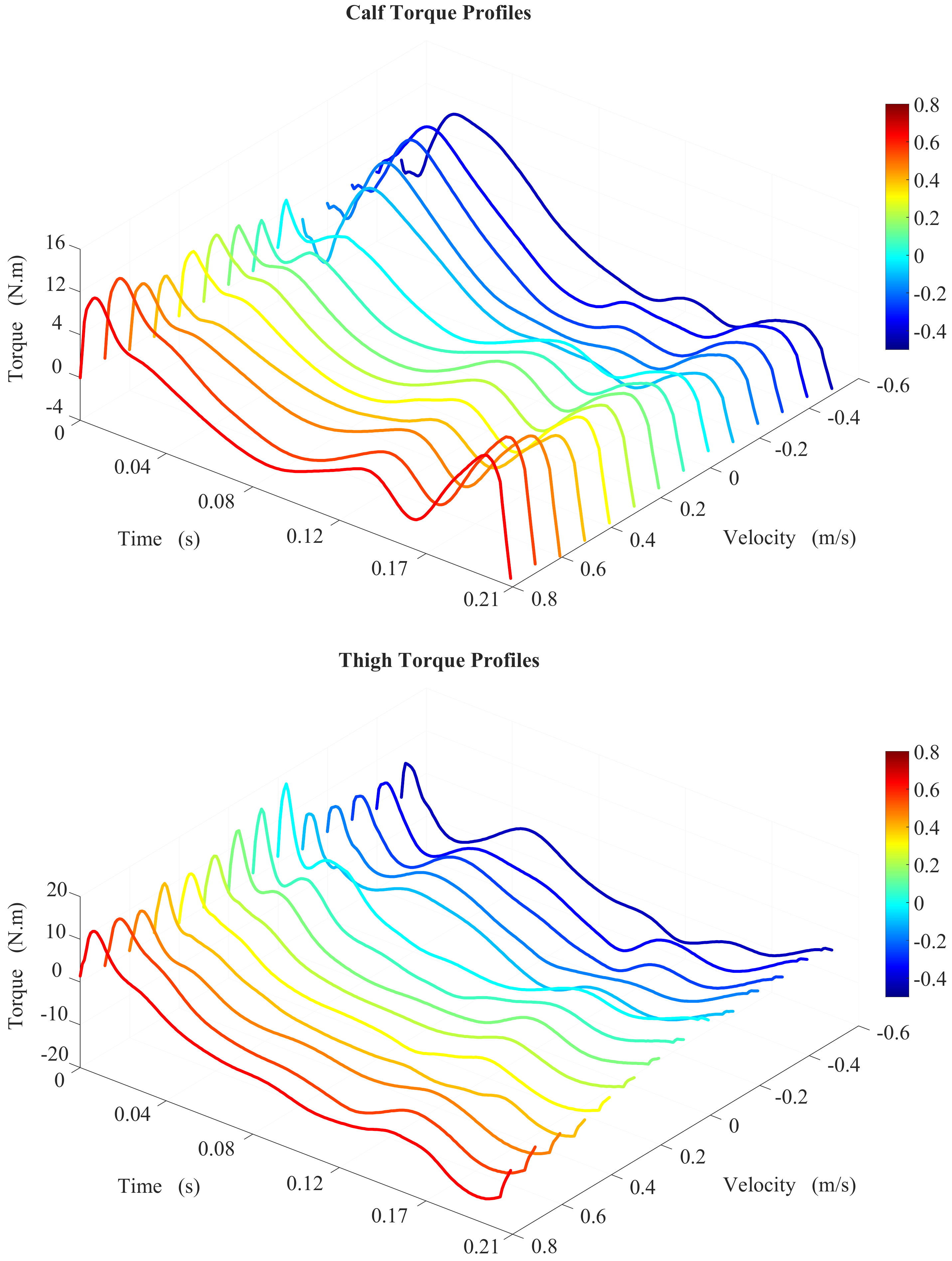}
\caption[Torque library]{This figure illustrates the torque profiles learned from the ILC. Once trajectories with varying average speeds in the GL are practiced, these torque profiles are converted into a look-up table and directly used for the online feedforward control in the proposed control as the Torque Library. 
}
\label{fig:Torques}
\end{figure}

\begin{table}
\caption{This table summarizes the improvements in the tracking performances for all solutions with varying average speeds from \SI[per-mode=symbol]{-0.5}{\meter\per\second} to \SI[per-mode=symbol]{0.8}{\meter\per\second} in the GL. Error columns show the error decrease (positive) or increase (negative) in percentage.}
\label{table:1}
\setlength{\tabcolsep}{0.5em} 
\renewcommand{\arraystretch}{1.1}
\begin{center}
\vspace{-4mm}
\begin{tabular}{|c||c|c|r||c|c|r|}
\hline 
\multirow{2}{1cm}{\centering{\textbf{Speed (m/s)}}} & \multicolumn{3}{c||}{ \textbf {Calf Angle (rad)}} & \multicolumn{3}{c|}{ \textbf {Thigh Angle (rad)}}  \\  \cline{2-7} 
& \it PD  & \it ILC & \it Error & \it PD & \it LC & \it Error \\ 
\hline
-0.5  & 0.123  &  0.032  &  73.7\%   &  0.047  &  0.016  &  66.6\%  \\  
-0.4  & 0.118  &  0.038  &  67.7\%   &  0.045  &  0.016  &  65.2\%   \\ 
-0.3  & 0.114  &  0.027  &  76.6\%   &  0.043  &  0.018  &  58.4\%  \\ 
-0.2  & 0.114  &  0.029  &  74.4\%   &  0.040  &  0.015  &  62.6\%  \\ 
-0.1  & 0.113  &  0.033  &  71.0\%   &  0.019  &  0.025  & -33.2\%  \\  
0.0   & 0.086  &  0.015  &  82.2\%   &  0.044  &  0.051  & -15.7\%  \\
0.1   & 0.083  &  0.012  &  85.2\%   &  0.032  &  0.043  & -36.3\%  \\
0.2   & 0.088  &  0.015  &  82.9\%   &  0.029  &  0.037  & -30.7\%  \\
0.3   & 0.089  &  0.015  &  83.6\%   &  0.034  &  0.033  &  2.9\%   \\
0.4   & 0.092  &  0.022  &  76.0\%   &  0.037  &  0.023  &  38.5\%  \\
0.5   & 0.089  &  0.019  &  79.0\%   &  0.039  &  0.019  &  52.1\%  \\
0.6   & 0.088  &  0.017  &  81.1\%   &  0.044  &  0.011  &  74.8\%  \\
0.7   & 0.090  &  0.038  &  57.4\%   &  0.040  &  0.010  &  75.5\%  \\
0.8   & 0.092  &  0.027  &  70.8\%   &  0.037  &  0.018  &  50.9\%  \\
\hline
\end{tabular}
\vspace{-6mm}
\end{center}
\end{table}
The resulting torque profiles are visualized in Fig.~\ref{fig:Torques} for calf and thigh joints of the front left leg accordingly.
All curves in these figures represent the averaged torques of $30$ successive strides collected from the stance phases in steady state.
In general, we find that the joint torques are slowly varying as the average speed changes.
The most noticeable changes happen during the beginning or the end of the stance phase.
When the robot is moving forward with a positive speed, the torques are ramping up rapidly in both of the calf and thigh joints at the beginning of the stride.
However, as speed increases, calf joint torques are increasing, while thigh joint torques are decreasing.
Near the end of the stance phase, the thigh joint torques become negative (clockwise direction) but similar trends are observed.
Additionally, we notice drastic changes in torques when the torso speed reverses its direction and the robot starts moving backward.
This may be caused by the leg configurations and the robot's asymmetrical design with respect to its frontal plane. 

\section{Conclusion and Discussion}
\label{sec:conclusion}

In this paper, we propose a control scheme using ILC for legged robots to rapidly learn unmodeled dynamics and to accurately achieve trajectory tracking without relying on heavy calculations or a large amount of robot data.
Once the ILC for a specific task is deployed, a robot can repetitively learn from its own mistakes from the previous trials and internalizes the synergies of motor commands as a lookup table linked to the specific task.
To some degree, this learning and memorizing process is strikingly similar to how animals in nature gain their ``muscle memories" \cite{KRAKAUER200658}.

Additionally, with the help of ILC, the values of feedback term $\vec{\tau}^{b}_{k}(s)$ in (\ref{eq:feedback}) are drastically reduced but better tracking performance is expected with much smaller feedback control gains.
It is because in the proposed control scheme, instead of generating desired torques to follow the trajectories, the feedback control only has to be used to reject the random disturbances from the environment.
This will improve the overall backdrivability of the system, which is crucial for legged robots to negotiate more complex terrain or to perform agile maneuvers. 

In the future work, we will teach the robot to learn more gait patterns such as trotting, bounding, and galloping.
Also, we will apply the proposed controller to train the robot to perform more tasks, such as ascending/descending stairs, sharp turning, and obstacle avoidance.
Furthermore, in order to improve the robustness of the learning control, a better state estimation is also required.
This can be implemented as an extended state observer similar to  \cite{9483268}.

\bibliographystyle{IEEEtran}
\bibliography{References}

\end{document}